\documentclass{article} 

\usepackage{iclr2027_conference,times}

\usepackage[utf8]{inputenc} 
\usepackage[T1]{fontenc}    
\usepackage[hidelinks]{hyperref} 

\usepackage{url}            
\usepackage{graphicx}       
\usepackage{booktabs}       
\usepackage{amsmath}        
\usepackage{amsfonts}       
\usepackage{nicefrac}       
\usepackage{microtype}      
\usepackage{xcolor}         
\usepackage{xspace}         
\usepackage{colortbl}       
\usepackage{pifont}         
\usepackage{amsthm}         
\usepackage[most]{tcolorbox} 
\usepackage{listings}       
\usepackage[acronym,nopostdot,nohypertypes={acronym}]{glossaries} 
\tcbset{
  myroundedbox/.style={
    colback=white,
    colframe=black,
    arc=3mm,
    boxrule=0.8pt,
    left=4mm, right=4mm, top=2mm, bottom=2mm,
  }
}

\newif\ifrevhighlight
\revhighlightfalse
\definecolor{revcolor}{RGB}{0,90,200}
\newcommand{\rev}[1]{\ifrevhighlight\textcolor{revcolor}{#1}\else#1\fi}

\newif\ifrawrhighlight
\rawrhighlightfalse
\definecolor{rawrcolor}{RGB}{190,30,140}
\newcommand{\rawrrev}[1]{\ifrawrhighlight\textcolor{rawrcolor}{#1}\else#1\fi}

\pdfmapline{+delphine < Poppins-SemiBold.ttf <T1-WGL4.enc}

\DeclareFontFamily{T1}{poppins}{}
\DeclareFontShape{T1}{poppins}{m}{n}{<->s*[1.0]Poppins-SemiBold}{}

\newcommand\delphinefont[1]{{\usefont{T1}{delphine}{m}{n} #1}}
\newcommand{\titlemethod}{\delphinefont{RAWR}\xspace}
\newcommand{\method}{{\fontsize{8.6pt}{11pt}\delphinefont{RAWR}}\xspace}

\newacronym{llm}{LLM}{Large Language Model}
\newacronym{prm}{PRM}{Process Reward Model}
\newacronym{ig}{IG}{Information Gain}
\newacronym{cot}{CoT}{Chain-of-Thought}
\newacronym[short={\method}]{mcnig}{\method}{Reward Assignment Without Rollouts}

\newcommand{\mcnigeight}{\method~8B\xspace}
\newcommand{\mcnigfourteen}{\method~14B\xspace}
\newcommand{\igeight}{IG~8B\xspace}

\newcommand{\outmult}{5}      
\newcommand{\costomega}{23}   
\newcommand{\costms}{77}      
\newcommand{\costqwen}{155}   
\newcommand{\rawomega}{4.8}   
\newcommand{\rawms}{15.7}
\newcommand{\rawqwen}{32}

\colorlet{ourrow}{green!8}  

\newcommand{\cmark}{\ding{51}}
\newcommand{\xmark}{\ding{55}}

\title{\titlemethod: Reward Assignment Without Rollouts in Verifiable Domains}

\author{
\makebox[\textwidth][c]{%
  \bfseries
  Corentin Royer$^{1}$ \qquad
  Anna Hedström$^{2}$ \qquad
  Debarun Bhattacharjya$^{3}$ \qquad
  Gaetano Rossiello$^{1}$
}\\[4pt]
\makebox[\textwidth][c]{%
  \bfseries
  Andrea Giovannini$^{1}$ \qquad
  Mennatallah El-Assady$^{4}$
}\\[6pt]
\makebox[\textwidth][c]{%
  \normalfont
  $^{1}$International Business Machines (IBM) \qquad
  $^{2}$ETH AI Center
}\\
\makebox[\textwidth][c]{%
  \normalfont
  $^{3}$Lirio \qquad
  $^{4}$Department of Computer Science, ETH Zurich
}
}

\iclrfinalcopy

\begin{document}

\maketitle

\begin{abstract}
Understanding and evaluating multi-step reasoning in \glspl{llm} at the level of individual steps remains a key challenge. \Glspl{prm} provide a solution by scoring each step, enabling fine-grained supervision and improved reliability. However, training them requires costly human annotation or computationally intensive rollout-based labeling. To solve this, we introduce \gls{mcnig}, a scalable method for automatically labeling the quality of individual reasoning steps in \rawrrev{any verifiable domain. Its step score, net information gain (NetIG),} improves upon single-reference \gls{ig} by comparing the most-supported correct answer against the most-supported incorrect one, yielding a robust signal even for long and structured outputs like code and SQL, where \gls{ig} fails. We show that the signal produced by \method\ correlates with human judgments of step quality, and we apply \method\ labels to train \glspl{prm} that achieve the best average best-of-$K$ accuracy across eight benchmarks spanning mathematics, code generation, text-to-SQL, and scientific QA. \rev{Crucially, \method\ generates no rollouts, cutting labeling complexity to $\mathcal{O}(N)$ and making it up to $\costms\times$ cheaper than rollout-based methods at comparable label quality, which makes large-scale process supervision practical.}
\end{abstract}

\glsresetall

\section{Introduction}

Reasoning models such as OpenAI o1~\citep{jaech2024openai} and DeepSeek-R1~\citep{guo2025deepseek} solve hard multi-step problems by generating long \gls{cot} traces and scaling test-time compute~\citep{snell2024scaling}. Yet this reasoning remains unreliable at the step level: in mathematics~\citep{hendrycks2021measuring}, code generation~\citep{zhuo2024bigcodebench}, text-to-SQL~\citep{li2024can}, and scientific or medical QA~\citep{xu2025ugphysics, jin2019pubmedqa, royer2024multimedeval}, a single flawed step can derail the solution, and even correct final answers often rest on erroneous intermediate steps~\citep{zheng2025processbench}.


Reward models (RMs) have been central to the development of conversational \glspl{llm}~\citep{ouyang2022training}, enabling alignment with human preferences and more reliable generation. They can be broadly categorized as outcome reward models (ORMs)~\citep{cobbe2021training, shi2022language} and \glspl{prm}~\citep{lightman2023let, math_shepherd, wang2024multi-mips, zhang2025lessons}. ORMs assign a single score to the final output, providing coarse-grained evaluation that suffices when only correctness matters. In contrast, a \gls{prm} takes a step-by-step answer to a question and grades each individual step. This fine-grained evaluation allows for more nuanced feedback and facilitates deeper inspection of where and how reasoning may go astray. By capturing the quality of individual steps, we improve the ability of the reward model to identify reasoning errors. This also enhances interpretability by identifying which steps contribute positively or negatively to the final answer. \Glspl{prm} are particularly useful in tasks requiring transparency and multi-step reasoning, such as programming and scientific QA.

Recent advances in test-time scaling and reinforcement learning have made research on \glspl{prm} more relevant than ever. A wide range of methods, from simple re-ranking and answer selection~\citep{wei2022chain, stiennon2020learning} to more sophisticated approaches like tree-based search~\citep{yao2023tree}, rely on accurate estimations of the value of individual reasoning steps, making them well-suited to benefit from the fine-grained supervision offered by \glspl{prm}, and when training using reinforcement learning, guiding the model with a \gls{prm} rather than an ORM leads to higher performance~\citep{math_shepherd}. Prior work has largely focused on mathematical reasoning tasks~\citep{lightman2023let, math_shepherd}, leaving domains such as code generation mostly unexplored. Expanding the range of tasks where a \gls{prm} can be effectively applied is crucial; unfortunately, gathering a large enough dataset to train such a model is a difficult and expensive endeavor, often requiring substantial human labeling effort. This hinders the scalability and adoption of \glspl{prm}, motivating the need for more efficient data labeling methods.

\begin{figure}[t]
  \centering
  \includegraphics[width=\linewidth]{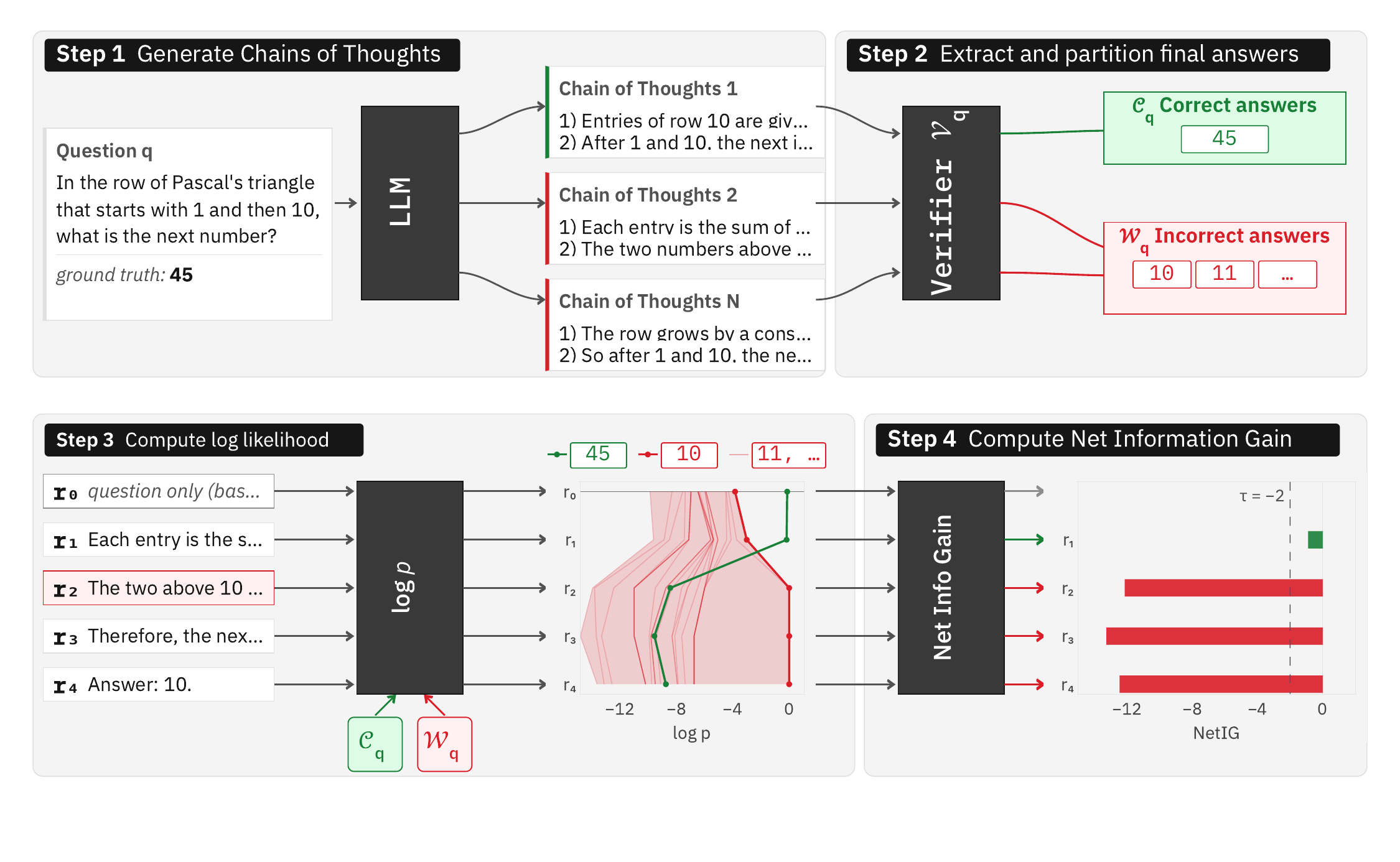}
  \vspace{-35pt}
  \caption{Overview of \method. \textbf{(1)}~An \gls{llm} samples $K$ \glspl{cot} for question $q$. \textbf{(2)}~A verifier $\mathcal{V}_q$ partitions them into correct ($\mathcal{C}_q$) and incorrect ($\mathcal{W}_q$) sets. \textbf{(3)}~At every prefix $r_0,\dots,r_n$, we score each candidate answer and keep the $\max$ over $\mathcal{C}_q$ and $\mathcal{W}_q$. \textbf{(4)}~The \rawrrev{step score, NetIG,} is the change in the correct-incorrect gap relative to $r_0$, thresholded at $\tau$ to produce binary step labels.}
  \label{fig:mcnig_example}
\end{figure}
\vspace{-5pt}

In this work, we introduce \gls{mcnig}, a method for automatically measuring the quality of individual reasoning steps in \gls{cot} traces. It requires only an \gls{llm} and a binary validator\rawrrev{, so it applies to any verifiable domain}: the \gls{llm} samples a set of solutions (\autoref{fig:mcnig_example}, step 1), and the validator partitions them into correct and incorrect answers (step 2), yielding a Monte Carlo approximation of the correct (\textit{resp.} incorrect) answer space. At every reasoning prefix, we compute the log-likelihood of each candidate answer (step 3) and measure how the step shifts the model toward the correct answer space relative to the incorrect one (step 4). \rawrrev{We call this signal the net information gain (NetIG) of the step, and threshold it into a binary reward. Unlike rollout-based methods, \method\ never samples continuations from intermediate prefixes: the $K$ solutions it draws are the traces being labeled, and the rest is a forward pass over them.}

The benefits of this contrastive formulation are many. First, it makes the labeling signal robust to tasks with long or compositional outputs, where simpler \gls{ig} measures degrade. Second, its technical requirements are minimal: \method\ is applicable to any autoregressive language model and any task equipped with a binary validator, it requires no human annotation, no generation beyond the sampled traces themselves, and no task-specific design choices beyond the validator itself. Another benefit of \method\ is its complexity: $\mathcal{O}(N)$ in the number of reasoning steps through KV-cache reuse, \rev{making it $\costomega\times$ cheaper to run than \textsc{OmegaPRM}~\citep{omegaprm}, the cheapest rollout-based method, once output tokens are priced at the commercial rate (\autoref{app:token_estimation})}, making it efficient and scalable. In this paper, we apply \method\ labels to train \glspl{prm}, but the step-quality scores are general-purpose to any application requiring fine-grained reasoning evaluation. We summarize our contributions as follows:

\begin{enumerate}
    \item \textbf{\rawrrev{Rollout-free reward assignment for reasoning steps.}} We introduce \method, a method for scoring individual \gls{cot} steps that achieves $\mathcal{O}(N)$ complexity.
    \item \textbf{Strong \glspl{prm} across diverse domains.} We apply \method\ to train \glspl{prm} that achieve the highest average best-of-$K$ accuracy across eight benchmarks spanning mathematics, code generation, text-to-SQL, and scientific QA.
    \item \textbf{Step labels that track human judgments.} \method\ scores increase monotonically with expert step labels on \texttt{PRM800K} and localize the first erroneous step in \texttt{ProcessBench} traces.
\end{enumerate}

More broadly, our work demonstrates that step-level supervision does not require rollouts: a single scoring pass over sampled reasoning traces is enough to train competitive \glspl{prm}, reducing the barrier from months of expert annotation or quadratic rollout budgets to a forward pass. By extending the scope of process supervision to code and SQL domains, we show that fine-grained step-level feedback benefits any task requiring multi-step reasoning, a setting common to most real-world applications of \glspl{llm}. Beyond \gls{prm} training, \method\ labels open avenues for reward shaping in reinforcement learning, and interpretability of multi-step reasoning.

\section{Related work}

\paragraph{Verifier-guided reasoning.}
\rev{Sampling multiple reasoning traces and selecting the best with a verifier improves \gls{llm} accuracy when single-shot accuracy is low~\citep{cobbe2021training}.} \citet{lightman2023let} scaled this to step-level verification by creating \texttt{PRM800K}, 800K expert-labeled reasoning steps for \texttt{MATH}~\citep{hendrycks2021measuring}; training a step classifier on it substantially improved best-of-$K$ selection and established \glspl{prm} as a tool for reasoning verification. \rev{The reliance on expert annotators, however, makes this expensive and hard to extend to new domains.}

\vspace{-5pt}
\paragraph{Reducing the cost of step-level labels.}
\rev{To avoid human annotators,} \textsc{MathShepherd}~\citep{math_shepherd} generates rollout completions from each reasoning prefix and measures the fraction reaching the correct answer, but requires $\mathcal{O}(N^2)$ token processing. \textsc{OmegaPRM}~\citep{omegaprm} reduces this to $\mathcal{O}(N \log N)$ with binary search over the first incorrect step, but still requires substantial computation. \rev{Our method achieves $\mathcal{O}(N)$ complexity (\autoref{app:complexity}) with no additional generations, only with a single scoring pass over existing reasoning traces. \autoref{tab:related_comparison} summarizes these methods.}

\begin{table}[ht]
  \caption{Comparison of step-level supervision methods. \method\ requires no rollouts and achieves $\mathcal{O}(N)$ complexity. Relevant prior methods have been validated on mathematics only.}
  \label{tab:related_comparison}
  \centering
  \resizebox{\linewidth}{!}{%
  \begin{tabular}{lcccc}
    \toprule
    Method & Label type & Complexity & Rollout-free & Cross-domain \\
    \midrule
    \texttt{PRM800K}~\citep{lightman2023let} & Human expert & --- & \cmark & \xmark \\
    \textsc{MathShepherd}~\citep{math_shepherd} & Rollout & $\mathcal{O}(N^2)$ & \xmark & \xmark \\
    \textsc{OmegaPRM}~\citep{omegaprm} & Rollout + binary search & $\mathcal{O}(N \log N)$ & \xmark & \xmark \\
    \textsc{QwenPRM}~\citep{zhang2025lessons} & Rollout + consensus filtering & $\mathcal{O}(N^2)$ & \xmark & \xmark \\
    \rowcolor{ourrow}\method\ (ours) & Information Theoretic & $\mathcal{O}(N)$ & \cmark & \cmark \\
    \bottomrule
  \end{tabular}}
\end{table}
\vspace{-5pt}

\vspace{-5pt}
\paragraph{Deriving process signals from outcome supervision.}

An alternative approach avoids step labels entirely by repurposing outcome-level supervision. \citet{yu2023ovm} introduce outcome-supervised value models that assess the potential of incomplete reasoning paths, reframing guided decoding as value estimation. \citet{yuan2024free} derive implicit process rewards from the log-likelihood ratio between a policy and a reference model, enabling \gls{prm}-like behavior from outcome-only training, and PRIME~\citep{cui2025process} uses this implicit \gls{prm} to provide dense rewards during reinforcement learning. While these methods eliminate step-level annotation, their step scores are implicit proxies learned from outcome labels rather than direct measures of step quality; in our experiments, OVM and ImplicitPRM trail \method\ by 5.7 and 11.5pp in average best-of-$K$ accuracy (\autoref{tab:prm_performance}).

\vspace{-5pt}
\paragraph{Information-theoretic analysis of reasoning.}
Recent work has applied information-theoretic tools to understand and evaluate \gls{cot} reasoning. \citet{chen2023rev} evaluate free-text rationales using conditional V-information, measuring the label-relevant content a rationale adds beyond the input. \citet{ton2024understanding} analyze information flow through reasoning traces, quantifying how intermediate steps influence the likelihood of correct answers. \citet{li2026making} introduce step entropy as a measure of redundancy in reasoning, showing that not all steps contribute equally to predictive power. These approaches use information measures primarily for \emph{analysis}, scoring existing rationales or diagnosing failures. \method\ instead operationalizes a contrastive \gls{ig} signal as a \emph{labeling} mechanism, directly producing training data for competitive \glspl{prm}.


\section{Preliminaries}
\label{sec:preliminaries}

We formalize the objects \method\ operates on (steps 1-2 of \autoref{fig:mcnig_example}): \gls{cot} traces sampled for a question $q$ and their partition into correct and incorrect answer sets $\mathcal{C}_q, \mathcal{W}_q$. We then review \gls{ig}, the single-reference predecessor of our signal, and identify its failure mode on long or structured answers, which \method\ is designed to fix.

\subsection{Problem setup and notation}
\label{sec:notation}

Let $p_\theta$ denote an autoregressive language model that defines a distribution over token sequences via $p_\theta(x_t \mid x_{<t})$. A \gls{cot} is a sequence of intermediate steps $R = (r_1, \ldots, r_N)$, where each step $r_i$ is a subsequence of tokens, and we write $R_{1:i} = (r_1, \ldots, r_i)$ for the prefix up to step $i$. The collection of traces for question $q$ is $\mathcal{R}_q = \{R_1, \ldots, R_K\}$.

From each \gls{cot} we extract a final answer $y = (y_1, \ldots, y_{|y|})$, which is a subsequence of the tokens of the last step, i.e.\ $\{y_1, \ldots, y_{|y|}\} \subseteq \{r_{N,1}, \ldots, r_{N,|r_N|}\}$, yielding a set of candidate answers $\mathcal{Y}_q = \{y_1, \ldots, y_K\}$. We write $y_{<t}$ for the first $t{-}1$ tokens of $y$. A domain-specific validator $V_q : \mathcal{Y} \to \{0,1\}$ determines whether a candidate answer is correct, and we partition:
\begin{equation}
\mathcal{C}_q = \{\, y \in \mathcal{Y}_q \mid V_q(y) = 1 \,\}, \qquad
\mathcal{W}_q = \{\, y \in \mathcal{Y}_q \mid V_q(y) = 0 \,\}.
\label{eq:answer_partition}
\end{equation}

\subsection{Information gain}
\label{sec:ig}

A natural approach to evaluating the quality of a reasoning step is to measure how much it increases the model's confidence in the correct answer~\citep{ton2024understanding}. Given a partial reasoning prefix $R_{1:i}$, we define the \emph{information} that $p_\theta$ assigns to an answer $y$ at step~$i$ as the total log-likelihood:
\begin{equation}
\mathcal{I}_i(y) = \sum_{t=1}^{|y|} \log p_\theta\!\left(y_t \,\middle|\, q,\, R_{1:i},\, y_{<t}\right).
\label{eq:information}
\end{equation}
When no reasoning is provided ($i{=}0$), this reduces to $\mathcal{I}_0(y) = \sum_{t=1}^{|y|} \log p_\theta(y_t \mid q, y_{<t})$. The \emph{\gls{ig}} at step~$i$ for a reference answer $y^\star$ is then:
\begin{equation}
\text{IG}_i = \mathcal{I}_i(y^\star) - \mathcal{I}_0(y^\star),
\label{eq:ig}
\end{equation}
measuring how much the partial reasoning $R_{1:i}$ improves the model's support for $y^\star$ relative to the no-reasoning baseline. We subtract the no-reasoning baseline $\mathcal{I}_0$ rather than the preceding step $\mathcal{I}_{i-1}$ because the cumulative reference is invariant to how the model partitions reasoning into steps: finer step granularity reduces per-step likelihood changes, which would make threshold-based labeling sensitive to formatting choices rather than reasoning quality.

\vspace{-5pt}
\paragraph{Limitations of single-reference \gls{ig}}
\label{sec:ig_limitations}

A critical limitation of \gls{ig} is its dependence on a \emph{single} reference formulation~$y^\star$. For mathematical tasks, this is benign: a numeric answer like ``$42$'' admits essentially one formulation. For structured outputs such as code or SQL, however, the space of semantically equivalent correct answers can grow rapidly with answer length. Variable naming choices (e.g., \texttt{result} vs.\ \texttt{res} vs.\ \texttt{ans}), equivalent expressions (e.g., \texttt{x != None} vs.\ \texttt{x is not None}), formatting conventions, and library alternatives each multiply the number of valid formulations. The number of valid formulations can thus grow combinatorially with answer length~$L$, though correlations between choices reduce it in practice; \gls{ig} degrades as it grows (\autoref{fig:labeling_and_shap}, right).



\section{Method}
\label{sec:mcnig}

\method\ replaces \gls{ig}'s single reference answer with a contrastive signal between the correct and incorrect answer sets (steps 3-4 of \autoref{fig:mcnig_example}). We motivate its two design choices (aggregation over formulations and correct-incorrect contrast) and give its full definition.

\subsection{Aggregating over answer formulations}
\label{sec:max_mode}

The limitation identified in \autoref{sec:ig_limitations} points to a natural target quantity: rather than measuring the model's support for a single reference answer, we want to estimate $\log p_\theta(\text{correct} \mid q, R_{1:i})$, the model's total log-probability of producing \emph{any} correct answer. Since the model assigns probability to individual token sequences rather than to equivalence classes, this quantity decomposes as:
\begin{equation}
\log p_\theta(\text{correct} \mid q, R_{1:i})
= \log \sum_{y \in \mathcal{C}_q} p_\theta(y \mid q, R_{1:i}).
\label{eq:logsumexp}
\end{equation}
Evaluating Eq.~\eqref{eq:logsumexp} exactly is intractable since $\mathcal{C}_q$ may be large, so we must estimate it from the finite set of $K$ sampled answers.

Note that we have access to individual log-likelihoods $\mathcal{I}_i(y) = \log p_\theta(y \mid q, R_{1:i})$ for each sampled answer $y$. We can use these sampled log-likelihoods to estimate the target quantity. However, this estimator accounts only for formulations that happen to appear among the $K$ samples, missing the tail of the equivalence class. For structured outputs such as code or SQL, where the number of valid formulations can be large, $K$ samples provide incomplete coverage and the estimator remains unstable as each newly discovered formulation increases the sum.

We instead propose to use the $\max$ operator: $\max_{y \in \hat{\mathcal{C}}_q} \mathcal{I}_i(y)$. This formulation has low sample complexity: it requires only that the sample contains \emph{one} instance of a high-probability formulation to stabilize, rather than covering the full support of the equivalence class. Formally, the $\max$ provides a lower bound on the target quantity:
\begin{equation}
\max_{y \in \hat{\mathcal{C}}_q} \mathcal{I}_i(y)
\;\leq\; \log \sum_{y \in \mathcal{C}_q} p_\theta(y \mid q, R_{1:i}),
\label{eq:max_bound}
\end{equation}
and the gap depends only on the concentration of $p_\theta$ around its preferred formulation, not on the size of the equivalence class. When the model concentrates most of its correct-answer mass on a small number of preferred phrasings, the $\max$ closely tracks the true class probability regardless of how many low-probability alternatives exist.
We validate this choice empirically in \autoref{app:ablations}: \rev{$\max$ outperforms summing over alternative answers (82.0\% vs.\ 81.6\% balanced accuracy).}

\subsection{Contrastive step scoring}
\label{sec:mcnig_def}

Building on the $\max$ estimator, we define the \emph{Net Information} at step~$i$ as the gap between the best-supported correct and incorrect answers:
\begin{equation}
\text{NetInfo}_i =
\underbrace{\max_{y \in \mathcal{C}_q} \mathcal{I}_i(y)}_{\substack{\text{most-supported}\\\text{correct answer}}}
\;-\;
\underbrace{\max_{y \in \mathcal{W}_q} \mathcal{I}_i(y)}_{\substack{\text{most-supported}\\\text{incorrect answer}}}.
\label{eq:netinfo}
\end{equation}

The \textbf{net information gain} (NetIG) at step~$i$, the score \method\ assigns to each step, is then the net information relative to baseline:
\begin{equation}
\text{NetIG}_i = \text{NetInfo}_i - \text{NetInfo}_0.
\label{eq:mcnig}
\end{equation}
\rev{By subtracting the pre-reasoning baseline, $\text{NetIG}_i$ measures the contribution of step $r_i$: positive values indicate the prefix shifts the model toward a correct outcome, while negative values indicate it introduces or reinforces errors. Anchoring at $r_0$ rather than $r_{i-1}$ captures reasoning that drifts into error gradually over several steps rather than at a single step (see \autoref{app:ablations} for the ablation). This matches how step-level supervision is usually defined: rollout methods condition on the prefix $R_{1:i}$ and human \texttt{PRM800K} annotations judge each step in the context of its predecessors.} \autoref{fig:mcnig_example} illustrates the full pipeline on a concrete example.

\subsection{Label assignment}

To train \gls{prm}s, we require binary labels. We thus project the continuous \rawrrev{NetIG} signal by thresholding:
\begin{equation}
\ell_i = \begin{cases} 1 & \text{if } \text{NetIG}_i > \tau_q, \\ 0 & \text{otherwise}, \end{cases}
\label{eq:labels}
\end{equation}
where $\tau_q$ is selected by optimizing balanced accuracy of trace-level labels (the product of step labels for each trace), ensuring the resulting binary labels are not biased toward positive or negative steps. \rev{Because the final answer $y$ is a subsequence of the last step $r_N$, we restrict the prefix sweep to $i < N$ to select the threshold, so the signal cannot degenerate by relying only on the last step.}


\section{Experiments}
\label{sec:experiments}

We evaluate \method\ along two complementary axes. First, we assess label quality directly by measuring how well \method\ scores separate correct from incorrect reasoning steps across domains with varying answer lengths (\autoref{sec:labeling_quality}). Second, we evaluate the labels indirectly through downstream \gls{prm} performance on best-of-$K$ selection across diverse benchmarks, including out-of-distribution settings (\autoref{sec:main_results}-\autoref{sec:generalization}).

\subsection{Experimental setup}
\label{sec:setup}

\vspace{-5pt}
\paragraph{Chain-of-thought generation.}
We prompt \textbf{Ministral-8B-Instruct-2410} with nucleus sampling ($T{=}1.0$, top-$p{=}0.95$) to generate $K{=}32$ completions per problem, formatted with \texttt{[STEP]} delimiters for consistent parsing. The mid-sized generator ensures a mix of correct and incorrect completions for contrastive scoring, and $K{=}32$ provides diverse formulations for the $\max$ operator \rev{(label accuracy plateaus by $K \approx 20$--$30$; \autoref{app:ablations})}. All 32 answers are retained for \method\ scoring. \rev{For \gls{prm} training, we remove unparseable answers, drop problems solved by all candidates, and subsample 8 traces per problem to control the correct/incorrect ratio, keeping a correct trace when available.} Answers are validated with domain-specific tools: \texttt{math-verify}~\citep{math_verify} for mathematics, unit-test execution for Python, and execution accuracy for SQL. \rev{The full scoring prompt template is given in \autoref{app:prompt_template}.} Full details are in \autoref{app:training_set}.

\vspace{-5pt}
\paragraph{Datasets.}
\rev{Our training set spans four domains (189K mathematics, 36K text-to-SQL, 5704 Python, and 1000 medical QA problems) in total, we have 232K samples and 1.49M candidate answers. We evaluate on eight benchmarks covering the same domains plus \texttt{AIME~23-25} and the out-of-distribution \texttt{UGPhysics}~\citep{xu2025ugphysics}. For the coding benchmarks we hold out a training subset (64 from \texttt{HumanEval}, 640 from \texttt{BigCodeBench}) and evaluate on the remainder.} All results use best-of-$K$ selection ($K{=}32$), where \gls{prm} scores are the product of step-level predictions and ORM scores the final classification probability. Full dataset details are in \autoref{app:training_set}.

\vspace{-5pt}
\paragraph{\Gls{prm} training.}
\rev{Given step-level labels, we train \glspl{prm} on the Ministral~3 8B and 14B backbones to predict correctness at each reasoning step. The input interleaves the question and steps with a delimiter token $\langle s_{\mathrm{req}} \rangle$:}
\[
x = \big(q,\; r_1,\; \langle s_{\mathrm{req}} \rangle,\; r_2,\; \langle s_{\mathrm{req}} \rangle,\; \ldots,\; r_N,\; \langle s_{\mathrm{req}} \rangle \big),
\]
\rev{and at each delimiter the model classifies the step via two reserved tokens $\langle s_{\texttt{POS}} \rangle, \langle s_{\texttt{NEG}} \rangle$, trained with binary cross-entropy averaged over steps (batch size 128, AdamW, learning rate $1\mathrm{e}{-6}$, max length 8{,}192, two epochs). ORM baselines are identical but classify only after the full trace.}

\subsection{Labeling quality and efficiency}
\label{sec:labeling_quality}

A central prediction of the analysis in \autoref{sec:ig_limitations} and~\autoref{sec:max_mode} is that single-reference \gls{ig} should degrade on tasks with long, structured answers, while \method\ should remain robust. We now validate this prediction empirically. 

\begin{figure}[t]
  \centering
  \includegraphics[width=\linewidth]{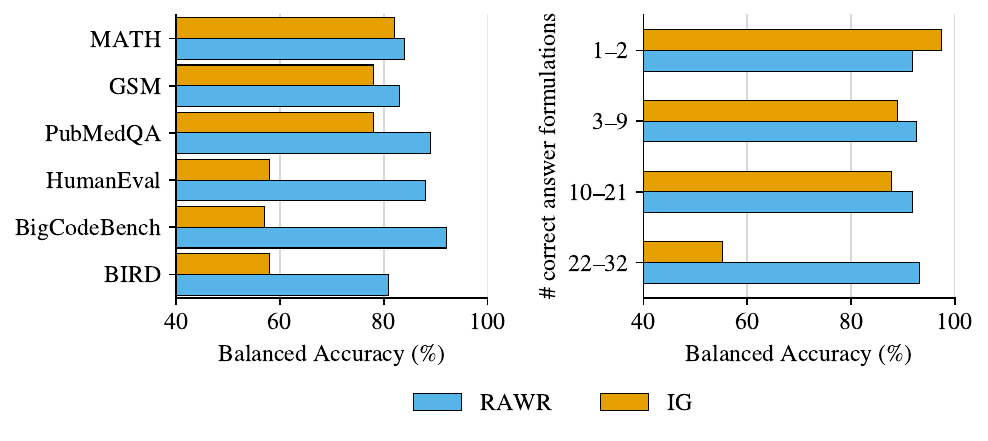}
  \vspace{-25pt}
  \caption{\textbf{Left:} Balanced accuracy of \gls{cot}-level labels for \gls{ig} vs.\ \method\ across domains. \textbf{Right:} Accuracy stratified by the number of distinct correct answer formulations. \Gls{ig} degrades as formulation diversity increases, while \method\ remains stable.}
  \label{fig:labeling_and_shap}
\end{figure}

\vspace{-5pt}
\paragraph{\method\ provides more reliable labels than \gls{ig}.}
We evaluate labeling quality by measuring the balanced accuracy of \gls{cot}-level predictions. Across all datasets, \method\ yields better accuracy than \gls{ig} (\autoref{fig:labeling_and_shap}, left). The improvement is especially pronounced on code generation (Python and SQL) tasks, precisely the domains where the theoretical analysis predicts \gls{ig} degradation due to long, compositional outputs.

\vspace{-5pt}
\paragraph{\method\ scores correlate with human judgments.}
To validate that \method\ captures genuine step quality rather than superficial statistical artifacts, we compare its scores against human annotations from \texttt{PRM800K}~\citep{lightman2023let}, which provides expert step-level labels (positive, neutral, negative) for mathematical reasoning traces. Mean \rawrrev{NetIG} scores increase monotonically across human label categories: $-0.38$, $+0.52$, and $+1.70$, confirming that the information-theoretic signal aligns with human judgments of reasoning quality. This is in line with previous findings that information gain correlates with human step-level evaluations~\citep{ton2024understanding}, and extends this validation to the \rawrrev{NetIG} estimator.

\vspace{-5pt}
\paragraph{Answer diversity explains the \gls{ig}-\method\ gap.}
To test whether answer diversity drives the \gls{ig}-\method\ gap, we stratify problems by the number of distinct correct formulations among the $K$ candidates (\autoref{fig:labeling_and_shap}, right). With 1--2 formulations (typical of math), \gls{ig} scores 97.4\% vs.\ 91.8\% for \method. At 22--32 formulations, \gls{ig} falls to 55.2\% while \method\ stays at 93.2\%, confirming that single-reference scoring breaks down when many equivalent answers exist and the $\max$ operator recovers the signal.

\vspace{-5pt}

\begin{figure}[t]
  \centering
  \includegraphics[width=0.85\linewidth]{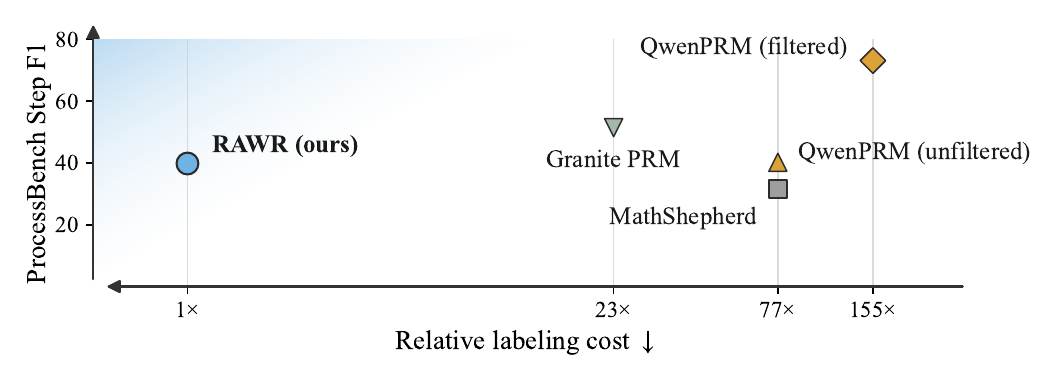}
  \vspace{-15pt}
  \caption{\rev{\texttt{ProcessBench} Step F1 against labeling cost relative to \method\ (\autoref{app:token_estimation}); top-left is best. Filtered QwenPRM adds an LLM-as-a-judge filtering stage to the same rollouts, raising F1 but discarding half of the labeled samples ($\costqwen\times$).}}
  \label{fig:efficiency_frontier}
\end{figure}

\vspace{-5pt}
\paragraph{Computational efficiency.}
\method\ needs a single teacher-forced scoring pass per trace with KV-cache reuse and \emph{generates nothing during labeling}: it scores the traces already sampled for training, so it adds none of the expensive output tokens rollouts produce. \rev{With output priced at $\outmult\times$ input, it is $\costomega$--$\costms\times$ cheaper than rollout-based labeling, and matches unfiltered QwenPRM and exceeds \textsc{MathShepherd} on \texttt{ProcessBench} Step F1 at $\costms\times$ lower cost (\autoref{fig:efficiency_frontier}, \autoref{app:token_estimation}).} This allows larger training sets for the same budget, and downstream \gls{prm} performance improves consistently with dataset size (\autoref{app:dataset_scaling}).

\subsection{\texorpdfstring{\Gls{prm}}{PRM} performance}
\label{sec:main_results}

\begin{figure*}[t]
  \centerline{\includegraphics[width=\linewidth]{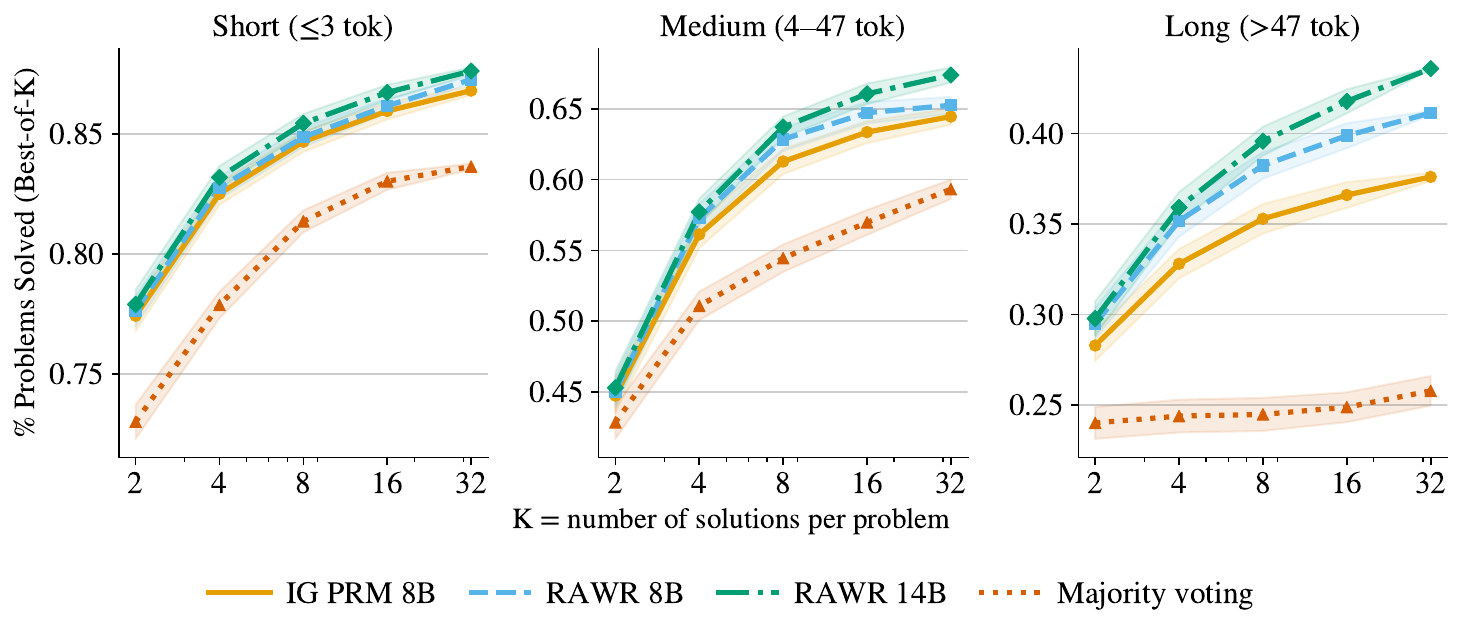}}
  \vspace{-10pt}
\caption{Best-of-$K$ accuracy as a function of $K$, stratified by ground-truth answer length\rawrrev{, for \glspl{prm} trained on \gls{ig} labels and on \method\ labels}. The \method-\gls{ig} gap is negligible for short answers and widens as answers grow longer, confirming that single-reference \gls{ig} degrades on structured, long-form outputs while \method\ remains robust. Shaded bands show $\pm 1$ standard deviation across repeated subsamples of $K$ answers per problem.}
\label{fig:best_of_n}
\end{figure*}
\vspace{-5pt}

\paragraph{Baselines.} We compare against methods spanning the full supervision spectrum (\autoref{tab:prm_performance}): no reward model (single sampling, majority voting); outcome-level supervision; implicit step signals derived from outcome training (ImplicitPRM~\citep{yuan2024free}, OVM~\citep{yu2023ovm}); and explicit process supervision (\textsc{MathShepherd}~\citep{math_shepherd}, Granite~PRM~\citep{granite2026mathprm}, and QwenPRM~7B~\citep{zhang2025lessons}). \rev{Granite~PRM uses \textsc{OmegaPRM}'s labeling method, but on responses from larger generators and with extra math and code fine-tuning.} Because these baselines use different backbones and training sets, we additionally train controlled variants (\mcnigeight, \mcnigfourteen, \igeight, ORM~8B) that share the Ministral~3 backbone and training data and differ only in the labeling signal (and size, for the 14B variant).

\vspace{-5pt}

\paragraph{Results.} \mcnigeight\ achieves 62.3\% average accuracy, the best among all baselines, and scaling to 14B (\mcnigfourteen) adds +1.1pp. \rev{This confirms the prediction from \autoref{sec:ig_limitations}: the \gls{ig}-\method\ gap is small on short-answer mathematics (+1.7pp on \texttt{MATH}, +0.5pp on \texttt{GSM8K}) but large on code and SQL with long, compositional answers (+3.8pp on \texttt{BigCodeBench}, +2.0pp on \texttt{BIRD}), while QwenPRM~7B is at its strongest on mathematics, its training domain, but lags on code and SQL. \autoref{fig:best_of_n} makes the length dependence explicit across all benchmarks: the \gls{ig}-\method\ gap is negligible for short answers ($\leq 3$ tokens) and largest for long ones ($>47$ tokens), exactly where many equivalent formulations exist.} \rev{A paired bootstrap over test problems (10K resamples, common problems) confirms \method\ significantly outperforms QwenPRM~7B ($p<0.001$) and \igeight\ ($p{=}0.025$); as \method\ and \gls{ig} differ only in the labeling signal, this isolates the contrastive term. The gap to our ORM baseline is within noise ($p{=}0.73$), an expected tie: the ORM is trained on the verifier's perfect final-answer labels, which best-of-$K$ scores directly, making it an upper bound on that task.}

\begin{table*}[t]
\caption{Best-of-$K$ accuracy ($K{=}32$) across eight benchmarks. Bold indicates the best result and underline the second best, computed separately within the 8B group and globally for \mcnigfourteen. Shaded rows are our methods; \mcnigeight achieves the highest average performance among 8B models.}
\label{tab:prm_performance}
\centering
\setlength{\tabcolsep}{4pt}\renewcommand{\arraystretch}{0.9}
\resizebox{\linewidth}{!}{%
\begin{tabular}{lrrrrrrrr}
\toprule
Method     & \texttt{MATH} & \texttt{GSM} & \texttt{PubMed} & \texttt{HumanEval} & \texttt{BigCode} & \texttt{BIRD} & \texttt{AIME} & Average \\
\midrule
Single sampling & 44.5\% & 73.3\% & 36.2\% & 55.5\% & 23.3\% & 19.6\% & 17.4\% & 38.5\% \\
Majority voting & 67.3\% & 91.2\% & 46.8\% & 66.1\% & 27.6\% & 37.6\% & 29.0\% & 52.2\% \\
OVM             & 62.8\% & 92.0\% & 59.4\% & 76.3\% & 34.9\% & 45.1\% & 26.1\% & 56.6\% \\
MathShepherd    & 56.1\% & 89.5\% & 49.7\% & 52.5\% & 29.5\% & 35.9\% & 23.0\% & 48.0\% \\
ImplicitPRM     & 59.3\% & 89.9\% & 44.3\% & 77.8\% & 29.3\% & 35.6\% & 19.6\% & 50.8\% \\
Granite PRM  & 62.5\% & 90.9\% & 49.8\% & 61.6\% & 31.4\% & 35.4\% & 27.0\% & 51.2\% \\
QwenPRM 7B      & \underline{74.3\%} & \textbf{94.0\%} & 54.1\% & 69.7\% & 31.5\% & 36.8\% & \textbf{37.0\%} & 56.8\% \\
\midrule
\rowcolor{ourrow}\igeight (ours)    & 73.1\% & 93.3\% & 49.0\% & \underline{84.5\%} & 43.0\% & 44.2\% & \underline{35.0\%} & 60.3\% \\
\rowcolor{ourrow}ORM 8B (ours)   & 73.8\% & 93.7\% & \textbf{62.0\%} & 84.0\% & \underline{44.1\%} & \textbf{47.8\%} & 29.0\% & \underline{62.1\%} \\
\rowcolor{ourrow}\mcnigeight (ours) & \textbf{74.8\%} & \underline{93.8\%} & \underline{57.7\%} & \textbf{84.6\%} & \textbf{46.8\%} & \underline{46.2\%} & 32.5\% & \textbf{62.3\%} \\
\midrule
\rowcolor{ourrow}\mcnigfourteen (ours) & \textbf{75.0\%} & \textbf{94.0\%} & \underline{59.9\%} & \textbf{85.6\%} & \textbf{47.1\%} & \textbf{50.1\%} & 32.0\% & \textbf{63.4\%} \\
\bottomrule
\end{tabular}}
\end{table*}
\vspace{-5pt}


\subsection{Out of distribution evaluation}
\label{sec:generalization}

\paragraph{Domain Shift.}
We evaluate on the AtomicPhysics subset of \texttt{UGPhysics}~\citep{xu2025ugphysics}, a physics domain absent from our training set. \mcnigeight achieves 14.7\% best-of-$K$ accuracy, best among 8B methods, and \mcnigfourteen reaches 15.1\%, nearly doubling single-sampling performance (\autoref{tab:generalization}).

\vspace{-5pt}
\paragraph{Chain-of-thought generator shift.}
\texttt{ProcessBench}~\citep{zheng2025processbench} evaluates step-level error localization in reasoning traces using Step F1 as its official metric. We additionally introduce Trace F1, which measures whether a trace is correctly classified as containing an error. These traces were generated by models unseen during training (Qwen, Llama). As reported in \autoref{tab:generalization}, \mcnigeight achieves 73.2 Trace F1, close to QwenPRM~7B (75.0) and substantially outperforming MathShepherd (41.1) and Granite~PRM (51.9). On Step F1, our models score lower (39.8 for \mcnigeight vs.\ 73.1 for QwenPRM~7B), indicating that \method\ labels carry more noise at the individual step level. \rev{In contrast, QwenPRM adds a separate LLM-as-a-judge consensus filtering stage on top of its rollouts. It reduces this noise but discards half of the labeled samples, doubling the labeling cost to $\costqwen\times$ that of \method\ (\autoref{app:token_estimation}); without it, their ProcessBench Step F1 drops to 40.2 at $\costms\times$ our cost, the same as \textsc{MathShepherd}.} Despite noisier step-level labels, \mcnigeight exceeds all non-QwenPRM baselines on Trace F1 (see \autoref{app:processbench_offset}).

\begin{table*}[ht]
\caption{External evaluation. \texttt{UGPhysics}: best-of-$K$ accuracy (OOD domain). \texttt{ProcessBench} Trace: F1 for classifying whether a trace contains an error (introduced by us). \texttt{ProcessBench} Step: official F1 for localizing the first incorrect step. Bold indicates the best result and underline the second best, computed separately within the 8B group and globally for \mcnigfourteen.}
\label{tab:generalization}
\centering
\setlength{\tabcolsep}{4pt}\renewcommand{\arraystretch}{0.9}
\resizebox{0.83\linewidth}{!}{%
\begin{tabular}{lccc}
\toprule
Method & \texttt{UGPhysics} & \texttt{ProcessBench} Trace & \texttt{ProcessBench} Step \\
\midrule
Single sampling & 8.0\% & --- & ---\\
Majority voting & 11.5\% & --- & ---\\
OVM             & 9.2\% &33.3  &18.6 \\
MathShepherd    & 10.0\% & 41.1 & 31.5\\
ImplicitPRM     & 10.6\% & 68.6 & 21.2\\
Granite PRM  & 11.0\% & 51.9 & \underline{51.4}\\
QwenPRM 7B      & 13.4\% & \textbf{75.0} & \textbf{73.1} \\
\midrule
\rowcolor{ourrow}\igeight (ours)    & \underline{13.7\%} & 70.1 & 35.5\\
\rowcolor{ourrow}ORM 8B (ours)   & 10.4\% & 72.8 & ---\\
\rowcolor{ourrow}\mcnigeight (ours) & \textbf{14.7\%} & \underline{73.2} & 39.8\\
\midrule
\rowcolor{ourrow}\mcnigfourteen (ours) & \textbf{15.1\%} & \textbf{75.1} & 40.8\\
\bottomrule
\end{tabular}}
\end{table*}
\vspace{-5pt}

\section{Conclusion}

\rev{\method\ replaces single-reference \gls{ig} with a contrastive log-likelihood signal: how much a step widens the gap between the best-supported correct and incorrect answers. It matches \gls{ig} on mathematics and pulls ahead on code and SQL, where many equivalent formulations exist (\autoref{fig:labeling_and_shap}), while generalizing well out of distribution (\autoref{tab:prm_performance}, \autoref{tab:generalization}). Its $\mathcal{O}(N)$ labeling complexity scales the method across four domains and cheaply into new ones.}

\vspace{-5pt}
\paragraph{Scope and limitations.}
\rev{\method\ makes three deliberate tradeoffs. First, it needs a binary validator, so it applies only to verifiable domains (mathematics, code, SQL), not open-ended tasks such as dialogue; where the validator is imperfect, label quality degrades gracefully, sub-linearly in its error rate (\autoref{app:ablations}). Second, it trades some per-step precision for throughput: without consensus filtering its step labels are noisier than QwenPRM's (the Step F1 gap, \autoref{tab:generalization}), whose filtered labels cost $\costqwen\times$ more to produce; the filtering stage can still be applied on top of \method\ labels when needed. The noise is absorbed at trace level, where our \glspl{prm} match or exceed all baselines on Trace F1 and best-of-$K$. Third, step-level validation against human annotations is possible only in mathematics, where step labels are available.}

\vspace{-5pt}
\paragraph{Outlook.}
\rev{Two extensions follow. Producing per-step scores without rollouts, \method\ can serve as a dense reward in RL pipelines at the cost of a forward pass; and since its only domain-specific component is the validator, it extends to multimodal and agentic settings where validators already exist, reducing the barrier to step-level supervision from months of expert annotation to a single scoring pass.}

\section*{Acknowledgments}

This research was supported by the ETH AI Center through an ETH AI Center postdoctoral fellowship to Anna Hedström.


\bibliographystyle{iclr2027_conference}
\bibliography{references}


\newpage
\appendix

\section{Complexity Derivation}
\label{app:complexity}

We analyze complexity in terms of \emph{tokens processed} by each method.  Let \(N\) be the number of \gls{cot} steps to label, \(\bar s\) the average tokens per step, \(T\) the average number of tokens in an answer \(y\), and \(S\) the number of sampled candidate answers used in the Monte Carlo approximation. Denote by \(|q|\) the number of tokens in the question. For comparison, let \(M\) be the number of rollouts per prefix used by other methods.  We count tokens processed by the language model (generation or scoring). We also assume standard KV-caching is available so that once a prefix is processed we do not reprocess its earlier tokens when scoring different completions.

\vspace{-5pt}
\paragraph{MathShepherd (naive rollouts).}
At each step \(i\) (for \(i=1,\dots,N\)) the naive method generates \(M\) rollouts continuing the prefix \(R_{1:i}\). Each rollout has expected length approximately \((N-i)\bar s\) tokens. Therefore the total expected number of rollout tokens is
\[
\mathbb{E}[\text{Tokens}_{\text{MS}}]
\;=\;
M\bar s \sum_{i=1}^{N}(N-i)
\;=\;
M\bar s\cdot\frac{N(N-1)}{2}.
\]
Since \(\frac{N(N-1)}{2}=\mathcal{O}(N^2)\), this method costs \(\mathcal{O}(M\bar s N^2)\)
tokens (i.e., \(O(N^2)\) in \(N\)).

\vspace{-5pt}
\paragraph{OmegaPRM (binary-search on first-failure).}
OmegaPRM reduces the number of \emph{checked prefixes} from $N$ to
$\lceil \log_2 N \rceil$: only $\mathcal{O}(\log N)$ indices are queried by
binary search. At each query, $M$ rollouts are generated to the point of
conclusion. If the location of the first incorrect step is assumed to be
uniformly distributed over the $N$ steps of the \gls{cot}, then the \emph{expected}
rollout length is $N/2$ steps.

Therefore, the expected number of tokens processed is
\[
\mathbb{E}[\text{Tokens}_{\Omega}]
\;=\;
M \bar s \, \tfrac{N}{2}\, \lceil \log_2 N \rceil
\]
Thus, the expected asymptotic complexity of OmegaPRM is
$\mathcal{O}(N\log N)$ tokens. This improves over the quadratic
cost of MathShepherd, though still grows faster in $N$ than the linear
complexity of our method.

\vspace{-5pt}
\paragraph{Our method: \gls{mcnig}}
To compute \(\text{NetInfo}_i\) for all \(i=1\ldots N\) we must evaluate
\(\mathcal{I}_i(y)\) for every sampled candidate \(y\in\mathcal{Y}\).  With
KV-caching the natural implementation is:

\begin{enumerate}
  \item Process the prompt \(q\) and the entire target \gls{cot} \(r_{1:N}\) once,
        storing the KV-cache and the boundary of each step \(i\).  This costs
        \(|q| + \sum_{i=1}^N |r_i| \approx |q| + N\bar s\) input tokens.
  \item For each step \(i\) and each sampled answer \(y\) compute the token
        log-probabilities of \(y\) conditioned on \(q, r_{1:i}\).  With KV
        caching this requires only processing the answer tokens, i.e.\ about
        \(T\) tokens per (prefix,answer) pair.
\end{enumerate}

Thus the scoring tokens are approximately \(N \times S \times T\).  Adding the
one-time prefix pass and the baseline \(i=0\) scores (which cost \(\approx S T\))
gives:
\[
\mathbb{E}[\text{Tokens}_{\text{\gls{mcnig}}}]
\;=\;
|q| + N\bar s + (N+1)\,S\,T
\]

If \(S\) and \(T\) are treated as constants independent of \(N\), the asymptotic complexity is in \(\mathcal{O}(N)\).  Thus \textbf{asymptotically in \(N\) and for constant \(S,T\) the \gls{mcnig} method has a linear complexity,} whereas MathShepherd has a quadratic complexity and OmegaPRM a complexity in \(\mathcal{O}(N\log N)\).

\subsection{Empirical token count estimation}
\label{app:token_estimation}

The token counts reported in \autoref{tab:token_processed} are estimated by instantiating the formulas below on the \texttt{PRM800K} dataset. For each trace we measure the number of reasoning steps $N$, the average tokens per step $\bar{s}$, the number of tokens in the question $|q|$, and the total tokens across the distinct candidate answers sampled for its question, $T_{\text{alt}}$. We set $M{=}8$ rollouts per prefix (following the original papers) and sum over all traces. \rev{All methods receive the same KV-cache assumption as the derivation above: a trace is prefilled once and its cache is branched at every step boundary, so a rollout or a scored answer only pays for its own new tokens.}

\rev{\paragraph{Input vs.\ output tokens.} A raw token count treats every token equally, but the two kinds of tokens a method processes have very different cost. \emph{Input} (prefill) tokens are read by the model in a single parallel pass; \emph{output} (decode) tokens are generated one at a time and are far more expensive, both in wall-clock throughput (decode is memory-bandwidth bound) and in price. \autoref{tab:api_pricing} lists the current API prices of Anthropic, OpenAI and Google, which together account for 88\% of enterprise LLM API usage~\citep{menlo2025state}: every listed model bills output at $5$--$8.3\times$ the input rate, with a median of $\outmult\times$~\citep{anthropic2026pricing,openai2026pricing,google2026pricing}. We therefore price output at $\outmult\times$ input for the cost-weighted totals. This distinction is decisive here, because the rollout-based methods spend almost their entire budget \emph{generating} rollouts (output), whereas \method\ performs only teacher-forced scoring---every token is read, none is generated, so its cost is $100\%$ input. \autoref{tab:token_processed} reports the two separately. The conclusion does not depend on the multiplier: counting raw tokens ($1\times$), rollout-based labeling still costs $\rawomega$--$\rawqwen\times$ more than \method.}

\begin{table}[!ht]
\caption{\rev{API list prices (USD per million tokens; standard tier, short context) of the three providers that together account for 88\% of enterprise LLM API usage~\citep{menlo2025state}, accessed 26 September 2026~\citep{anthropic2026pricing,openai2026pricing,google2026pricing}. Every model bills output at $5$--$8.3\times$ the input rate (median $\outmult\times$), the multiplier used in \autoref{tab:token_processed}.}}
\label{tab:api_pricing}
\centering
\begin{tabular}{llrrr}
\toprule
Provider (API share) & Model & Input & Output & Output / input \\
\midrule
Anthropic (40\%) & Claude Fable 5.1 & 10.00 & 50.00 & $5.0\times$ \\
                 & Claude Opus 5.5 & 4.00 & 20.00 & $5.0\times$ \\
                 & Claude Sonnet 5 & 2.00 & 10.00 & $5.0\times$ \\
                 & Claude Haiku 4.5 & 1.00 & 5.00 & $5.0\times$ \\
\midrule
OpenAI (27\%)    & GPT-6 Astra & 10.00 & 50.00 & $5.0\times$ \\
                 & GPT-6 Sol & 2.00 & 10.00 & $5.0\times$ \\
                 & GPT-6 Luna & 0.10 & 0.50 & $5.0\times$ \\
\midrule
Google (21\%)    & Gemini 3.1 Pro (preview) & 2.00 & 12.00 & $6.0\times$ \\
                 & Gemini 3.8 Flash & 0.75 & 3.75 & $5.0\times$ \\
                 & Gemini 3.5 Flash-Lite & 0.30 & 2.50 & $8.3\times$ \\
\bottomrule
\end{tabular}
\end{table}

\begin{table}[!ht]
\caption{\rev{Tokens processed by each labeling method to produce the same number of labeled samples on \texttt{PRM800K}, split into input (prefill/scoring) and output (generation), after normalizing by each method's retention rate $\rho$. \Gls{cot} generation is shared across all methods and excluded. Rollout-based methods spend $94$--$98\%$ of their tokens on output; \method\ generates none. The cost-weighted total prices output at $\outmult\times$ input, the median list-price ratio of the major API providers (\autoref{tab:api_pricing}); the last two rows give the cost relative to \method\ at that rate and, for reference, counting raw tokens ($1\times$). \textsc{QwenPRM} uses the \textsc{MathShepherd} rollouts plus a separate LLM-as-a-judge filtering stage; its judge verdict tokens are excluded, so its cost is a lower bound. Granite~PRM uses Automatic Process Supervision (the \textsc{OmegaPRM} MCTS labeling method), so we estimate its labeling cost as equal to \textsc{OmegaPRM}'s; this counts the labeling stage only, not its additional fine-tuning.}}
\label{tab:token_processed}
\centering
\resizebox{\linewidth}{!}{%
\begin{tabular}{lccccc}
\toprule
& Qwen PRM & MathShepherd & OmegaPRM & Granite PRM & \cellcolor{ourrow}\textbf{\gls{mcnig}} \\
\midrule
Input tokens        & $1.8 \times 10^8$ & $0.44 \times 10^8$ & $0.44 \times 10^8$ & $0.44 \times 10^8$ & \cellcolor{ourrow}$\mathbf{1.5 \times 10^8}$ \\
Output tokens       & $45.8 \times 10^8$ & $22.9 \times 10^8$ & $6.6 \times 10^8$ & $6.6 \times 10^8$ & \cellcolor{ourrow}$\mathbf{0}$ \\
\midrule
Cost-weighted (out $=\outmult\times$) & $230.6 \times 10^8$ & $114.9 \times 10^8$ & $33.6 \times 10^8$ & $33.6 \times 10^8$ & \cellcolor{ourrow}$\mathbf{1.5 \times 10^8}$ \\
\quad relative to \method & $\costqwen\times$ & $\costms\times$ & $\costomega\times$ & $\costomega\times$ & \cellcolor{ourrow}$\mathbf{1\times}$ \\
\quad relative to \method, raw tokens (out $=1\times$) & $\rawqwen\times$ & $\rawms\times$ & $\rawomega\times$ & $\rawomega\times$ & \cellcolor{ourrow}$\mathbf{1\times}$ \\
\bottomrule
\end{tabular}}
\end{table}

Because each method discards a different fraction of samples before \gls{prm} training (e.g., samples where all rollouts are correct or all are incorrect), the raw token counts do not reflect the cost of reaching the same final dataset size. To ensure a fair comparison, we normalize every method by its \emph{retention rate} $\rho$---the fraction of labeled samples that survive filtering---so that the reported count reflects the total tokens that must be processed to produce the same number of usable training examples:
\begin{equation}
\text{Tokens}_{\text{adjusted}} = \frac{\text{Tokens}_{\text{raw}}}{\rho}.
\end{equation}
\textsc{MathShepherd} and \textsc{OmegaPRM} report no filtering stage ($\rho{=}1$), \textsc{QwenPRM} retains approximately 50\% of samples after consensus filtering ($\rho_{\text{Qwen}} = 0.50$), and \method\ retains approximately 83\% ($\rho_{\text{\gls{mcnig}}} = 0.83$).

The per-trace formulas, split into input and output tokens, are:

\paragraph{\textsc{MathShepherd}.} The trace is read once; from each prefix $R_{1:i}$, $M$ rollouts continue for the remaining $(N-i)$ steps:
\[
\text{In}_{\text{MS}} = |q| + N\bar{s}, \qquad
\text{Out}_{\text{MS}} = M \bar{s}\, \frac{N(N-1)}{2}.
\]

\paragraph{\textsc{OmegaPRM}.} Binary search checks $\lceil \log_2 N \rceil$ prefixes, each with $M$ rollouts of expected length $N/2$ steps:
\[
\text{In}_{\Omega} = |q| + N\bar{s}, \qquad
\text{Out}_{\Omega} = M \bar{s}\, \frac{N}{2}\, \lceil \log_2 N \rceil.
\]

\paragraph{\textsc{QwenPRM}.} The \textsc{MathShepherd} rollouts, plus a separate LLM-as-a-judge filtering stage that reads the question and the full trace:
\[
\text{In}_{\text{Qwen}} = \underbrace{|q| + N\bar{s}}_{\text{rollouts}} + \underbrace{|q| + N\bar{s}}_{\text{judge}}, \qquad
\text{Out}_{\text{Qwen}} = M \bar{s}\, \frac{N(N-1)}{2}.
\]
The judge's own verdict tokens are unknown and excluded, so this is a \emph{lower bound} on the filtering cost. Most of the filtering cost comes from the retention rate: discarding half of the labeled samples doubles the tokens needed per usable sample.

\paragraph{\method.} A single pass over the trace, plus scoring every candidate answer after each prefix (including the baseline $i{=}0$); nothing is generated:
\[
\text{In}_{\text{\gls{mcnig}}} = |q| + N\bar{s} + (N + 1)\, T_{\text{alt}}, \qquad
\text{Out}_{\text{\gls{mcnig}}} = 0.
\]

\clearpage
\section{Training Set Details}
\label{app:training_set}

To construct a diverse and comprehensive training corpus, we aggregate a wide range of publicly available benchmarks spanning mathematical reasoning, scientific question answering, program synthesis, and text-to-SQL semantic parsing. \autoref{tab:training_data} summarizes the datasets used, along with the number of unique problems and total candidate answers derived from each source.

For some coding benchmarks, we adapt a subset of the available samples for training to preserve held-out data for evaluation. Specifically, we use 64 samples from \texttt{HumanEval} and 640 samples from \texttt{BigCodeBench} as training data, reserving the remainder for testing. For \texttt{SynSQL}, we subsample 20,000 problems from the ``highly complex'' subset of the dataset for training.

\begin{figure}[!ht]
  \vskip 0.2in
  \centering
    \includegraphics[width=0.6\linewidth]{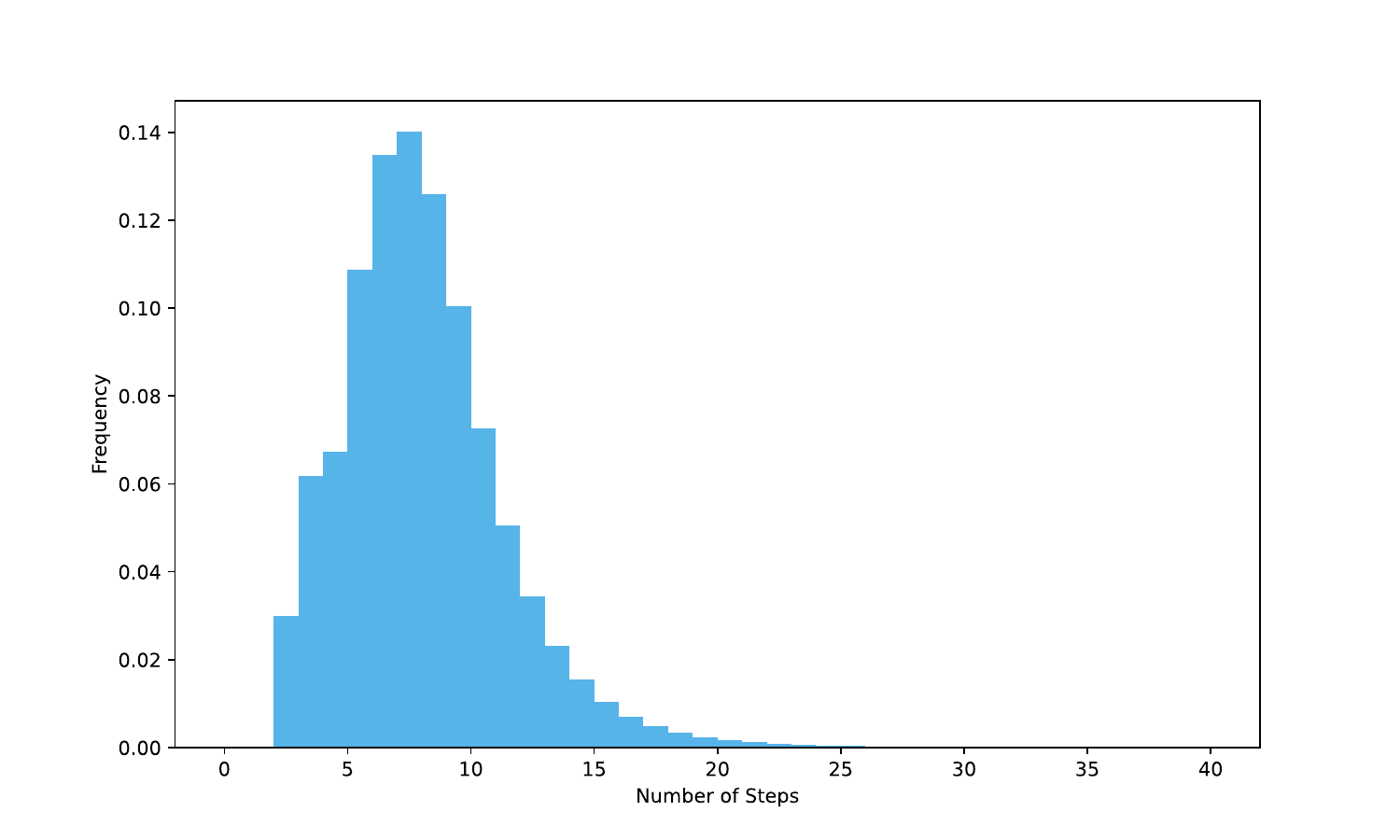}
    \caption{Distribution of the number of steps in the \glspl{cot} of all the datasets combined.}
    \label{fig:step_histo}
\end{figure}

Overall, the combined training mixture spans \textbf{232K problems} and, after generation and filtering, results in \textbf{1.49M candidate answers}, providing broad coverage of reasoning tasks and ensuring varied supervision across mathematical, scientific, and programming domains. \autoref{tab:cot_stats} reports statistics of the generated \glspl{cot}, and \autoref{fig:step_histo} shows the distribution of \gls{cot} lengths.

\begin{table}[!ht]
\centering
\caption{Training dataset composition.}
\label{tab:training_data}
\begin{tabular}{lrr}
\toprule
Dataset & \# Problems & \# Candidate answers \\
\midrule
\texttt{MATH500}~\citep{lightman2023let}         & 12{,}000   & 85{,}248 \\
\texttt{GSM8K}~\citep{cobbe2021training}          & 8{,}790    & 31{,}768 \\
\texttt{MathQA}~\citep{amini2019mathqa}           & 29{,}837   & 204{,}974 \\
\texttt{AquaRat}~\citep{ling2017program}          & 97{,}700   & 589{,}446 \\
\texttt{NumGLUE}~\citep{mishra2022numglue}        & 41{,}018   & 265{,}532 \\
\texttt{PubMedQA}~\citep{jin2019pubmedqa}         & 1{,}000    & 3{,}727 \\
\texttt{HumanEval}~\citep{chen2021evaluating}     & 64         & 480 \\
\texttt{BigCodeBench}~\citep{zhuo2024bigcodebench}& 640        & 5{,}080 \\
\texttt{APPS}~\citep{hendrycksapps2021}           & 5{,}000    & 26{,}768 \\
\texttt{BIRD}~\citep{li2024can}                   & 9{,}428    & 69{,}760 \\
\texttt{Spider}~\citep{yu2018spider}              & 7{,}000    & 46{,}233 \\
\texttt{ARCHER}~\citep{zheng-etal-2024-archer}    & 414        & 3{,}304 \\
\texttt{SynSQL}~\citep{li2025omnisql}             & 20{,}000   & 159{,}104 \\
\midrule
\textbf{Total} & \textbf{232{,}891} & \textbf{1{,}491{,}424} \\
\bottomrule
\end{tabular}
\end{table}

\begin{table*}[!ht]
    \caption{Statistics of the generated CoTs. The BoK columns show the oracle best-of-$K$: the percentage of problems where at least one of the $K$ samples is correct, whereas the \% correct columns show the percentage of generations that are correct.}
    \label{tab:cot_stats}
    \centering
    \begin{tabular}{llllll}
    \toprule
         & & \multicolumn{2}{l}{Train}    & \multicolumn{2}{l}{Test} \\
        Dataset & Avg \# step & BoK & \% correct & BoK & \% correct \\
    \midrule
        \texttt{MATH500} &8.50 & 86.57\% & 33.37\% & 88.98\% & 51.10\% \\
        \texttt{GSM8K} &7.52 & 98.55\% & 75.81\% & 97.72\% & 84.30\% \\
        \texttt{MathQA} &8.08 & 81.22\% & 38.52\% & - & - \\
        \texttt{AquaRat}& 8.29 & 71.65\% & 28.19\% & - & - \\
        \texttt{NumGLUE} & 5.80& 86.05\% & 38.34\% & - & - \\
        \texttt{PubMedQA} & 10.08& 87.35\% & 24.58\% & 86.97\% & 44.25\% \\
        \texttt{HumanEval} & 7.28& 98.44\% & 31.64\% & 95.96\% & 60.01\% \\
        \texttt{BigCodeBench} &8.07 & 69.12\% & 11.82\% & 68.07\% & 27.04\% \\
        \texttt{APPS} & 8.01& 51.41\% & 12.71\% & - & -  \\
        \texttt{BIRD} &7.25 & 67.11\% & 17.71\% & 74.49\% & 34.81\% \\
        \texttt{Spider} & 6.46& 90.07\% & 44.81\% & - & - \\
        \texttt{ARCHER}& 7.69 & 42.86\% & 6.31\% & - & - \\
        \texttt{SynSQL} & 7.48& 36.54\% & 6.70\% & - & - \\
        \texttt{UGPhysics} & 7.92 & - & - & 26.99\% & 9.11\% \\
        \texttt{AIME}& 11.57 & - & - & 54.00\% & 18.70\%\\
        \bottomrule
    \end{tabular}
\end{table*}

\clearpage
\section{Dataset Scaling}
\label{app:dataset_scaling}

\begin{figure}[!ht]
  \vskip 0.2in
  \centering
    \includegraphics[width=0.6\linewidth]{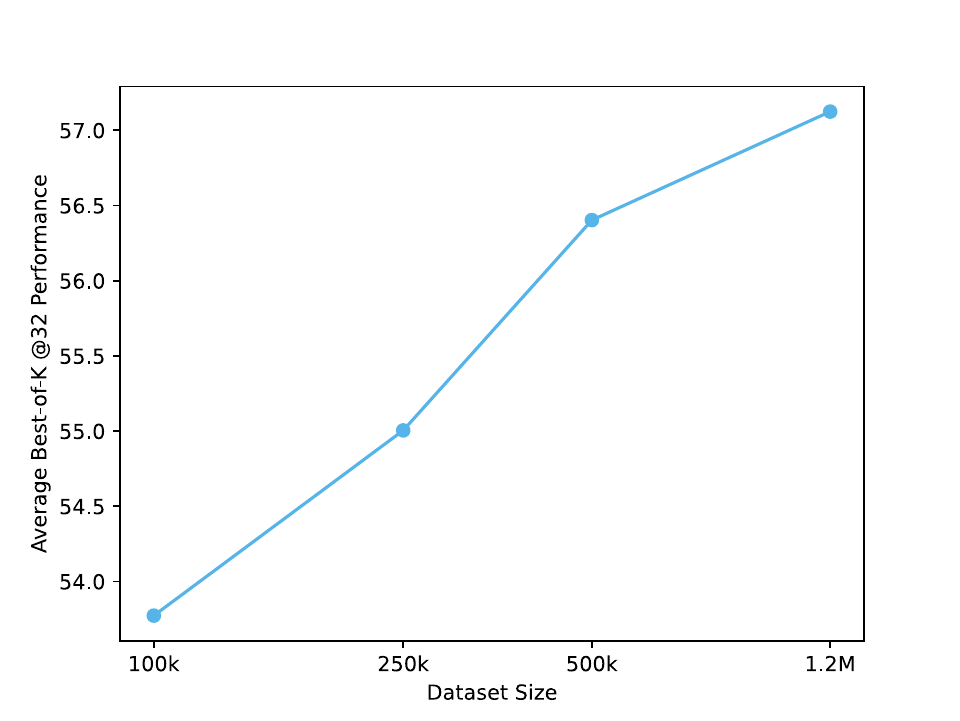}
    \caption{Average best-of-$K$ performance over the evaluation tasks for \glspl{prm} trained on different subsets of the training set, ranging from 100K to the full 1.2M samples.}
    \label{fig:dataset_scaling}
\end{figure}

We investigate how the scale of the \gls{prm} training dataset affects model performance. Using the Ministral~3 8B backbone, we train \glspl{prm} on four dataset sizes: 1.2M, 500K, 250K, and 100K samples. We then evaluate each resulting model on the full suite of best-of-$K$ tasks and report the average. As shown in \autoref{fig:dataset_scaling}, performance improves consistently with increased data scale, indicating that larger preference-label datasets provide more robust learning signals for reward modeling. These results highlight the value of \method\ as it allows for easier scaling up to millions of samples.

\clearpage
\section{\method\ Ablations}
\label{app:ablations}

\rev{We ablate two design choices: the aggregation rule over alternative answers ($\max$ vs.\ sum) and the baseline anchor. \method\ anchors each step against the pre-reasoning baseline $\text{NetInfo}_0$ (\autoref{eq:mcnig}); an alternative is the strictly local increment $\text{NetInfo}_i - \text{NetInfo}_{i-1}$, which measures only the change contributed by step $r_i$. \autoref{tab:anchor_ablation} reports balanced accuracy for all four combinations, micro-averaged across the 13 training datasets. $\max$ outperforms sum under both anchors (by $0.4$pp with the step-$0$ anchor), and anchoring at step~$0$ is uniformly better (e.g.\ $82.0\%$ vs.\ $80.5\%$ for $\max$). Reasoning traces frequently drift into error gradually, with no single step dropping the answer likelihood sharply; the successive difference cancels this accumulated drift along with the prior, whereas anchoring at step~$0$ retains it. This is why we keep the step-$0$ anchor rather than reformulate \method\ as a strictly marginal signal.}

\begin{table}[!ht]
    \caption{\rev{Balanced accuracy (\%) of \method\ labels under each aggregation rule ($\max$, sum) and baseline anchor, micro-averaged across the 13 training datasets. $\max$ outperforms sum under both anchors, and anchoring at the pre-reasoning baseline ($\text{NetInfo}_0$) outperforms the strictly local successive difference ($\text{NetInfo}_{i-1}$) under every aggregation rule.}}
    \label{tab:anchor_ablation}
    \centering
    \begin{tabular}{lcc}
    \toprule
        Aggregation & Baseline anchor ($r_0$) & Successive diff.\ ($r_{i-1}$) \\
    \midrule
        \texttt{max}  & \textbf{82.01} & 80.46 \\
        \texttt{sum}  & 81.60 & 80.09 \\
    \bottomrule
    \end{tabular}
\end{table}

\rev{\paragraph{Likelihood evolution along the trace.} \method\ rests on the assumption that a correct reasoning prefix raises the model's likelihood of the gold answer while a flawed one lowers it. We test this directly on \texttt{ProcessBench} (\texttt{GSM8K} and \texttt{MATH}, 1{,}241 traces), the shortest-answer datasets available (1-3 answer tokens) and thus the strongest case for a saturated prior. Under teacher forcing we measure the length-normalized probability of the gold answer $p_i(y^\star)$ given the prefix $R_{1:i}$, binned by normalized step index $i/N$ (\autoref{tab:likelihood_evolution}). Before any reasoning ($i/N=0$) the two groups are indistinguishable ($0.149$ vs.\ $0.140$); as reasoning accumulates they separate monotonically in opposite directions, ending at $0.995$ for correct traces and $0.027$ for incorrect ones---a $0.97$ gap. This is the behaviour the signal assumes, in the regime least favourable to it.}

\begin{table}[!ht]
    \caption{\rev{Median length-normalized gold-answer probability $p_i(y^\star)$ against the normalized step index $i/N$, teacher-forced on 1{,}241 \texttt{ProcessBench} traces. Correct and incorrect traces are indistinguishable before reasoning and diverge monotonically as it accumulates.}}
    \label{tab:likelihood_evolution}
    \centering
    \begin{tabular}{lcccccc}
    \toprule
        $i/N$ & $0.0$ & $0.2$ & $0.4$ & $0.6$ & $0.8$ & $1.0$ \\
    \midrule
        Correct ($n{=}484$)   & 0.149 & 0.153 & 0.165 & 0.376 & 0.886 & 0.995 \\
        Incorrect ($n{=}757$) & 0.140 & 0.133 & 0.113 & 0.107 & 0.069 & 0.027 \\
    \bottomrule
    \end{tabular}
\end{table}

\rev{\paragraph{Prior saturation and calibration.} A likelihood-based signal could lose its range if the model is already confident in the answer before any reasoning. We find this regime is rare and does not hurt label quality. First, saturation is rare even in the shortest-answer regime: the median pre-reasoning prior $p_0$ is $0.15$ on \texttt{ProcessBench}, no \texttt{GSM8K} problem and only $1.4\%$ of \texttt{MATH} problems exceed $p_0 > 0.9$, and the same holds at scale ($1.5\%$ of our full 85{,}248-trace \texttt{MATH} set). Second, where it does occur the labels hold: on the saturated subset, balanced accuracy is $89.9\% \pm 0.8$, above the $83.0\% \pm 0.1$ over the full set (\autoref{tab:prior_saturation}). \method\ thresholds a \emph{relative} difference, so it requires only that the score order correct above flawed reasoning, not that any probability be calibrated; a badly calibrated model can still order steps correctly.}

\begin{table}[!ht]
    \caption{\rev{Balanced accuracy of \method\ labels on the full \texttt{MATH} training set versus the high-prior subset ($p_0 > 0.9$), same protocol as the main paper. Spreads are $\pm 1$ std over a bootstrap. Labels on saturated priors are more accurate than on the full set.}}
    \label{tab:prior_saturation}
    \centering
    \begin{tabular}{lcc}
    \toprule
         & Balanced acc. & Traces \\
    \midrule
        Full set                 & $83.0\% \pm 0.1$ & 85{,}248 \\
        High-prior ($p_0 > 0.9$) & $89.9\% \pm 0.8$ & 1{,}298 \\
    \bottomrule
    \end{tabular}
\end{table}

\rev{\paragraph{Sensitivity to verifier noise.} \method\ depends on a binary validator to partition candidates. Our validators are execution- or parser-based and near-ground-truth, but we quantify sensitivity to an imperfect one by corrupting the partition: at noise level $\varepsilon$ we independently flip each candidate's label with probability $\varepsilon$ and recompute \method\ labels under the corrupted partition. \autoref{tab:verifier_noise} reports balanced accuracy on \texttt{BIRD} and \texttt{MATH} over 7 seeds per level. Degradation is sub-linear in $\varepsilon$: a $5\%$ error rate costs $3.1$ points on \texttt{BIRD} and $2.3$ on \texttt{MATH}, and even at $10\%$ the labels retain $80.2\%$ and $76.9\%$. This follows from the $\max$ aggregation---a flipped label moves the signal only if it becomes the argmax of the set it is wrongly placed in, so a misclassified low-probability answer is absorbed with no effect.}

\begin{table}[!ht]
    \caption{\rev{Balanced accuracy of \method\ labels under a verifier that mislabels each candidate with probability $\varepsilon$ (7 seeds per level). Degradation is sub-linear and the labels remain usable across the full range tested.}}
    \label{tab:verifier_noise}
    \centering
    \begin{tabular}{lcccc}
    \toprule
        Verifier noise $\varepsilon$ & \texttt{BIRD} & $\Delta$ vs.\ clean & \texttt{MATH} & $\Delta$ vs.\ clean \\
    \midrule
        $0.00$ (clean) & $85.34\%$ & --- & $81.59\%$ & --- \\
        $0.01$ & $84.41\%$ & $-0.93$ & $81.50\%$ & $-0.09$ \\
        $0.05$ & $82.22\%$ & $-3.13$ & $79.30\%$ & $-2.29$ \\
        $0.10$ & $80.15\%$ & $-5.19$ & $76.94\%$ & $-4.65$ \\
    \bottomrule
    \end{tabular}
\end{table}

\rev{\paragraph{Sampling budget $K$.} \method\ approximates the correct and incorrect answer spaces from $K$ sampled candidates. We ablate this budget on 100 \texttt{MATH} problems, drawing a 512-candidate reference pool and measuring downstream balanced accuracy when only $s$ candidates are subsampled (\autoref{tab:sampling_budget}). Accuracy rises steeply and plateaus by $s \approx 20$-$30$, staying within $4$ points of the full-pool reference. Our $K{=}32$ therefore sits on the plateau, and the method degrades gracefully below it ($s{=}10$, under a third of the budget, stays within 5.2 points of the reference). This follows from the low sample complexity of the $\max$ aggregator, which needs only one high-probability correct formulation to appear rather than full coverage of the equivalence class.}

\begin{table}[!ht]
    \caption{\rev{Downstream balanced accuracy as a function of the number of subsampled candidates $s$ (100 \texttt{MATH} problems, 512-candidate reference pool). Accuracy plateaus by $s \approx 20$-$30$; the paper's $K{=}32$ sits on the plateau.}}
    \label{tab:sampling_budget}
    \centering
    \begin{tabular}{lcccccc}
    \toprule
        $s$ & 5 & 10 & 20 & 30 & 40 & 512 (ref) \\
    \midrule
        Balanced acc.\    & 69.8\% & 72.7\% & 73.9\% & 74.2\% & 74.1\% & 77.9\% \\
        Gap to reference  & 8.1 & 5.2 & 4.0 & 3.7 & 3.8 & --- \\
    \bottomrule
    \end{tabular}
\end{table}

\begin{figure}[!ht]
    \centering
      \vskip 0.2in
      \includegraphics[width=0.4\linewidth]{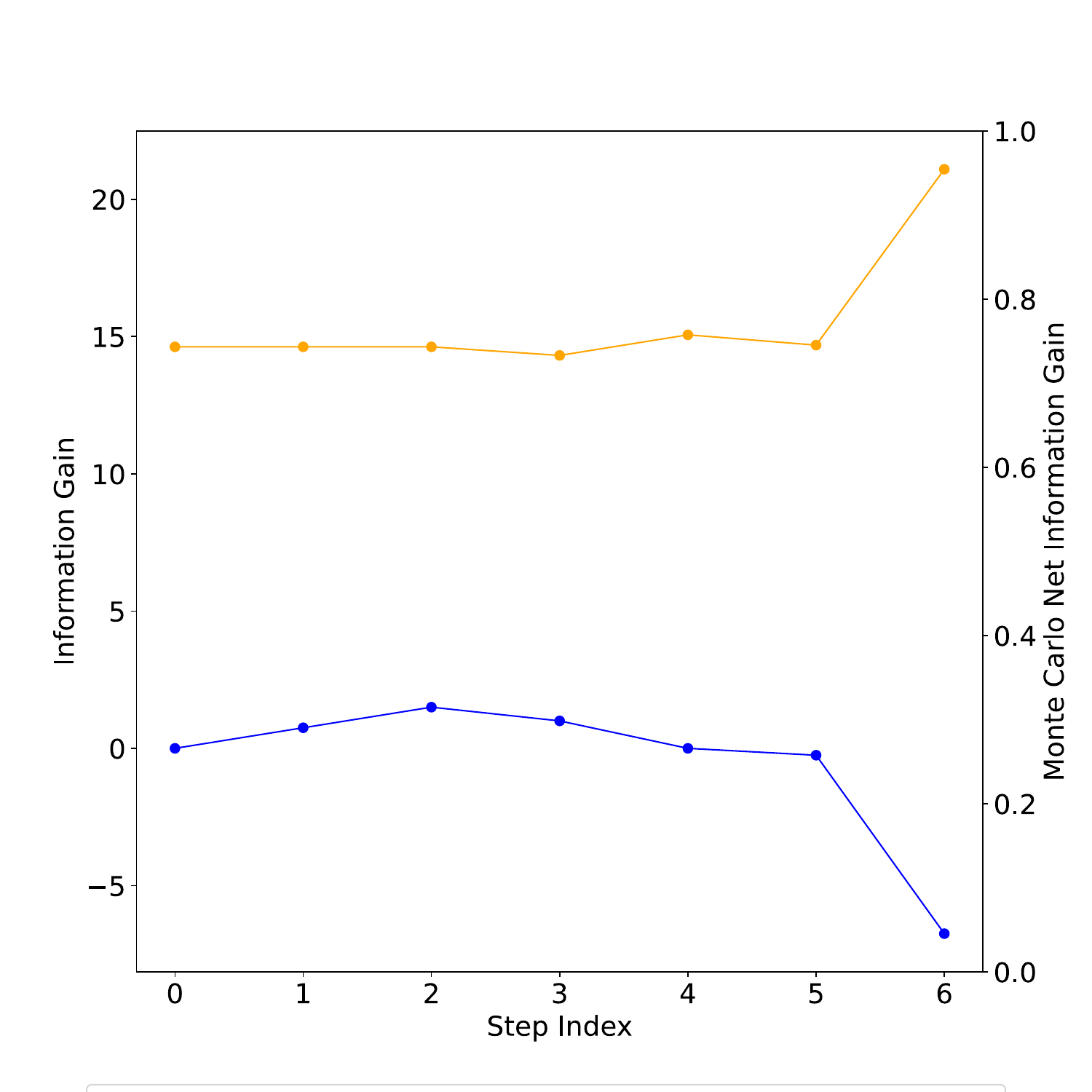}
    \caption{Example of a sample that is misclassified under \gls{ig} but correctly identified under \method, illustrating the sharper discriminative behavior of the max-based information signal.\rawrrev{ The right axis shows the NetIG score of \method.}}
    \label{fig:ig_vs_mcnig_example}
\end{figure}

Finally, we demonstrate the higher performance of \method\ compared to \gls{ig} in \autoref{fig:ig_vs_mcnig_example}. This shows the two signals for the same sample. The \gls{cot} reaches a correct answer but not the gold one from the dataset. As we approach the last step, the likelihood of generating the answer that \gls{ig} expects decreases whereas \method\ still performs well.

\vspace{.3in}

\begin{tcolorbox}[myroundedbox, title=Example of a \textit{correct} \gls{cot} for the Bird dataset (database schema omitted)]
\textbf{Question:} Indicate the title of all podcasts in the \texttt{fiction} category.
\begin{enumerate}
    \item Identify podcasts labeled as \texttt{fiction}.
    \item The \texttt{podcasts} and \texttt{categories} tables contain podcast metadata and category assignments, respectively.
    \item Join the tables on \texttt{podcast\_id} to combine podcast information with category information.
    \item Filter rows where the category equals \texttt{'fiction'}.
    \item Select unique podcast titles. The resulting SQL query is:
    \begin{lstlisting}[]
SELECT DISTINCT podcasts.title
FROM podcasts
INNER JOIN categories
    ON podcasts.podcast_id = categories.podcast_id
WHERE categories.category = 'fiction';
    \end{lstlisting}
\end{enumerate}
\end{tcolorbox}

\clearpage
\section{ProcessBench Error Localization Analysis}
\label{app:processbench_offset}

To better understand the Step F1 gap between \method\ and QwenPRM on \texttt{ProcessBench}, we analyze the distribution of \emph{error step offsets}: the difference between the predicted error step and the ground-truth error step. An offset of 0 indicates exact localization; negative offsets mean the \gls{prm} flags an error earlier than ground truth, and positive offsets mean it flags it later.

\begin{figure}[!ht]
    \centering
    \includegraphics[width=0.48\linewidth]{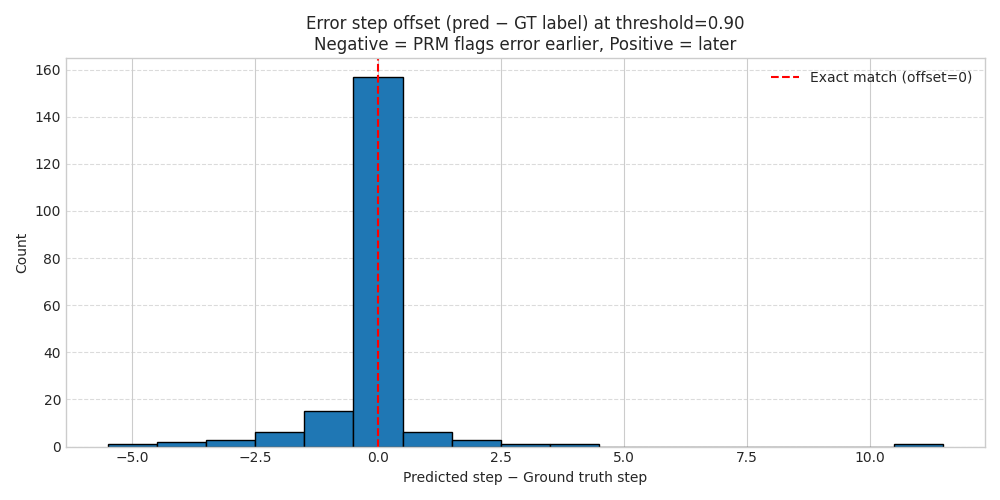}
    \hfill
    \includegraphics[width=0.48\linewidth]{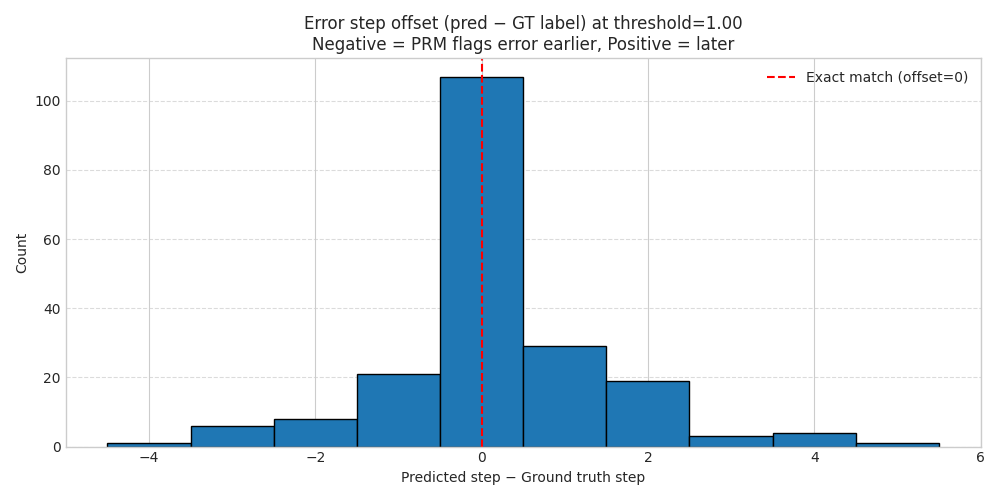}
    \caption{Error step offset distributions on \texttt{ProcessBench}. \textbf{Left:} QwenPRM~7B, tightly concentrated at offset~0, indicating precise step-level localization. \textbf{Right:} \mcnigeight, with the mode still at offset~0 but wider spread predominantly in $[-2, +2]$. The right skew (about 56 traces at positive offsets vs.\ 36 at negative ones) indicates a tendency to flag errors later than ground truth.}
    \label{fig:error_offset}
\end{figure}

\autoref{fig:error_offset} (left) shows the offset distribution for QwenPRM~7B. The distribution is tightly concentrated at offset~0 ($\sim$158 exact matches), with minimal spread, consistent with its high Step F1 of 73.1. \autoref{fig:error_offset} (right) shows the same analysis for \mcnigeight. The mode is still at offset~0 ($\sim$107 exact matches), but the distribution is wider, with notable mass at offsets $\pm 1$ and $\pm 2$.

\rev{A further check distinguishes \method\ from an outcome-level signal, which would concentrate its flags at the final step of a trace. \autoref{tab:before_last} reports, among incorrect \texttt{ProcessBench} solutions that \method\ flags, the fraction whose first flagged step falls \emph{before} the last step. Across all four subsets this exceeds $77\%$, confirming that the signal localizes errors inside the reasoning rather than defaulting to the outcome.}

\begin{table}[!ht]
    \caption{\rev{Percentage of flagged incorrect \texttt{ProcessBench} solutions whose first flagged step precedes the final step. An outcome-level signal would place its flag at the last step, giving values near $0$.}}
    \label{tab:before_last}
    \centering
    \begin{tabular}{cccc}
    \toprule
        \texttt{GSM8K} & \texttt{MATH} & \texttt{Olympiad} & \texttt{OmniMath} \\
    \midrule
        82.8\% & 77.2\% & 81.5\% & 80.2\% \\
    \bottomrule
    \end{tabular}
\end{table}

\clearpage
\section{Scoring Prompt Template}
\label{app:prompt_template}

\rev{\method\ scores each candidate answer by teacher-forcing it after a reasoning prefix. The generator and the scoring pass share the prompt below: a system instruction to reason in \texttt{[STEP]}-delimited steps and to close with the answer inside \texttt{<final-answer>} tags, four few-shot exemplars in that format, then the target question. To score a prefix, we truncate the trace after step $i$, append the candidate answer, and compute $\mathcal{I}_i(y)$ on the answer tokens only (those following \texttt{\textbackslash boxed\{} for mathematics, or inside \texttt{<final-answer>} for the other domains). The exemplars are drawn from the corresponding training domain; we show the mathematics template as a representative instance.}

\begin{tcolorbox}[myroundedbox, title=\rev{\method\ scoring prompt (mathematics; exemplar bodies abbreviated)}]
\rev{\textbf{System.} You are a careful reasoner. Solve the problem one step at a time. Begin each step with the marker \texttt{[STEP]}. After the final step, give the final answer between \texttt{<final-answer>} and \texttt{</final-answer>} tags, with the numeric result in \texttt{\textbackslash boxed\{\}}.}

\rev{\textbf{Few-shot exemplars} ($\times 4$, one shown):}

\rev{\emph{Question:} \dots}

\rev{\texttt{[STEP]} \dots\ \texttt{[STEP]} \dots}

\rev{\texttt{<final-answer>}\texttt{\textbackslash boxed\{}\dots\texttt{\}}\texttt{</final-answer>}}

\rev{\textbf{Target.}}

\rev{\emph{Question:} \{question\}}
\end{tcolorbox}


\end{document}